\def\year{2019}\relax
\documentclass[letterpaper]{article} %DO NOT CHANGE THIS
\usepackage{aaai19}  %Required
\usepackage{times}  %Required
\usepackage{helvet}  %Required
\usepackage{courier}  %Required
\usepackage{url}  %Required
\usepackage{graphicx}  %Required

\usepackage{algorithm}
\usepackage{algorithmic}
\usepackage{amsmath}
\usepackage{bm}
\usepackage{multirow}
\usepackage{diagbox}
\usepackage{arydshln}
\usepackage{pifont}
\usepackage{subfigure}

\begin{document}
\title{Adversarial Noise Layer: \\Regularize Neural Network By Adding Noise
}
\author{
  Zhonghui You\textsuperscript\dag, Jinmian Ye\textsuperscript\ddag, Kunming Li\textsuperscript\S, Zenglin Xu\textsuperscript\ddag, Ping Wang\textsuperscript\dag \\
  \textsuperscript\dag School of Electronics Engineering and Computer Science, Peking University\\
  \textsuperscript\ddag University of Electronic Science and Technology of China\\
  \textsuperscript\S Australian National University\\
  zhonghui@pku.edu.cn, jinmian.y@gmail.com, u5580030@anu.edu.au \\ zenglin@gmail.com, pwang@pku.edu.cn
}
\maketitle
\begin{abstract}
In this paper, we introduce a novel regularization method called Adversarial Noise Layer (ANL) and its efficient version called Class Adversarial Noise Layer (CANL), which are able to significantly improve CNN's generalization ability by adding carefully crafted noise into the intermediate layer activations. ANL and CANL can be easily implemented and integrated with most of the mainstream CNN-based models. We compared the effects of the different types of noise and visually demonstrate that our proposed adversarial noise instruct CNN models to learn to extract cleaner feature maps, which further reduce the risk of over-fitting. We also conclude that models trained with ANL or CANL are more robust to the adversarial examples generated by FGSM than the traditional adversarial training approaches.
\end{abstract}

\noindent Although Convolutional Neural Networks (CNNs) are powerful and widely used in various computer vision tasks, they suffer from over-fitting due to the excessive amount of parameters~\cite{srivastava2014dropout}. The initial development of the neural network was inspired by the mechanism of human brain~\cite{wang2017origin} which does not work as precisely as the computer. Inspired by the difference, we infer that adding noise into the process of training could instruct CNNs to learn more robust feature representations to against the effect of noise, thereby reducing the risk of over-fitting.

Many regularization methods~\cite{zhong2017random,simonyan2014very,krizhevsky2012imagenet} have been proposed to prevent over-fitting by adding noise into the training data. Besides, methods like DisturbLabel~\cite{xie2016disturblabel} randomly changed the label of a small subset of samples to incorrect value each iteration, thereby regularizing the CNNs on loss layer.

\citeauthor{Sankaranarayanan18} proposed a regularization method called Layerwise Adversarial Training (LAT)~\cite{Sankaranarayanan18} which uses the gradients of the previous batch to generate noise for the current batch during training. Different from data augment methods, LAT adds the perturbations not only to the input images, but also the intermediate layer activations.

In this work, we propose a variant of LAT and its efficient version, we call them Adversarial Noise Layer (ANL) and Class Adversarial Noise Layer (CANL) respectively. Unlike LAT, ANL generates the noise base on the current batch gradients. CANL further reduces additional time costs by caching the gradients based on image category.

%In this work, we further propose two well compatible and applicable regularization method, which improves the robustness of CNNs obviously by injecting the well-designed noise into the network in training. We name the noise which is calculated based on the gradient as the adversarial noise since it's similar to the perturbation calculation in Adversarial Training~\cite{goodfellow2014explaining}.

%Differ from Adversarial Training (AT), Adversarial Noise Layers (ANLs) add the noise to the input images as well as the hidden layers. \uline{Different from LAT .... }

ANL and CANL are well compatible with various CNN architectures and can be injected without changing the design philosophy. ANL and CANL are only embedded in the model during the training, so that it actually takes no extra computation in the inference process. Our main contributions are summarized as follows:

\begin{itemize}
\item We introduce a CNN regularization method called ANL. The empirical results show that ANL can significantly improve the performance of various mainstream deep convolution neural networks on popular datasets (Fashion-MNIST, CIFAR-10, CIFAR-100, ImageNet).
\item We introduce the CANL algorithm, which is an efficient version of ANL. Compared to ANL, CANL takes less training time, but at the cost of slight degeneration in regularization performance.
\item We demonstrate that ANL and CANL provide stronger regularization compared to LAT and Dropout. We also verified that ANL and CANL can improve the robustness of the CNN models, comparable to the traditional adversarial training approach, under the attack of Fast Gradient Sign Method~\cite{goodfellow2014explaining}.
\end{itemize}

%We verified that ANL and CANL can improve the robustness of the CNN model under the attack of the fast gradient symbol method, which is equivalent to the traditional confrontation training method.

\section{Related Work}

The recent rapid progress of CNNs in computer vision tasks such as image classification, semantic segmentation~\cite{DBLP:conf/iccv/HeGDG17} and image restoration~\cite{DBLP:journals/corr/abs-1711-10098} is partly due to the creation of large-scale dataset such as ImageNet~\cite{DBLP:journals/ijcv/RussakovskyDSKS15}\textit{ etc.} It is commonly believed that robust models usually require large-scale sufficient training data set or expensive computation to avoid over-fitting~\cite{DBLP:conf/aistats/LeeXGZT15}. Therefore, to overcome the over-fitting challenge, various methods and techniques are proposed such as optimization techniques~\cite{srivastava2014dropout}, regularization methods and\textit{ etc.}

The early solutions for avoiding over-fitting is to constrain the parameters by using the $\ell_2$ regularization~\cite{krizhevsky2009learning} or to stop training before convergence. Various regularization methods aim to reduce over-fitting are proposed recently. With a certain probability, the dropout method reduces over-fitting in large feed-forward neural networks by masking a random subset of the hidden neurons with zero during training~\cite{srivastava2014dropout} but performs unsatisfactorily in tiny or compact networks. DropConnect~\cite{wan2013regularization} randomly masks the weights with zero in training phase. Stochastic Pooling~\cite{zeiler2013stochastic} changes the deterministic pooling operation with randomly selects activations according to a certain distribution during training.

Recently, regularization methods by adding noise are introduced~\cite{krizhevsky2012imagenet,simonyan2014very,zhong2017random}. Simonyan\textit{ et al.}~\cite{simonyan2014very} introduced the methods to avoid over-fitting by random horizontal flipping train data, which directly enlarge the training dataset. Zhong\textit{ et al.}~\cite{zhong2017random} and Krizhevsky\textit{ et al.}~\cite{krizhevsky2012imagenet} further proposed data augmentation methods through erasing training and crop training dataset randomly. Adversarial Training~\cite{goodfellow2014explaining} enlarged training data set by adding adversarial training data to reduce the model sensitivity and improve the robustness of CNN models. DisturbLabel~\cite{xie2016disturblabel} added noise at the loss layer through randomly changing the label of a small subset of samples to incorrect values during training, thereby regularizing the CNN models. 

Different from the methods mentioned above, our approaches regularizes CNN models by adding adversarial noise in hidden layers, learning more robust feature representations. ANL can be integrated easily into the most CNN-based models and obtain better performance.

\section{Method}

\subsection{Terminology and Notation}
To simplify, in this paper we use the following notations and terminologies to illustrate our algorithm:

\begin{itemize}
	\item $J(\bm x, y; \bm \theta)$ denotes the cost function used to train the model, where $\bm x$ denotes the input image, $y$ denotes the corresponding true label, and $\bm \theta$ denotes the parameters of the model.
	\item $\bm h_t$ denotes the output of $t^{th}$ layer of the neural network.
	\item $s(\bm h_t)$ is the standard deviation of $\bm h_t$. 
    \item $\bm \eta_t$ is the adversarial noise for $\bm h_t$. Suppose the network has $L$ layers, and we treat the input as the $0^{th}$ layer, $\bm \eta = (\bm \eta_0, \bm \eta_1, \bm \eta_2,... ,\bm \eta_L)^T$ denotes the entirety of the adversarial noise.
    \item $\epsilon$ is the hyper-parameter used to control the magnitude of the noise for ANL and CANL.
    \item $Clip_{<a, b>}(\bm v)$ denotes the element-wise clipping $\bm v$, with $\bm v_i$ clipped to the range $[a,b]$.
	\item $\mathcal N(\mu, \sigma^2)$ denotes the Gaussian distribution where $\mu$ is the mean and $\sigma$ is the standard deviation (std).
\end{itemize}

\subsection{Adversarial Noise Layer}
\begin{figure*}[t]
  \centering
  \includegraphics[width=0.9\textwidth]{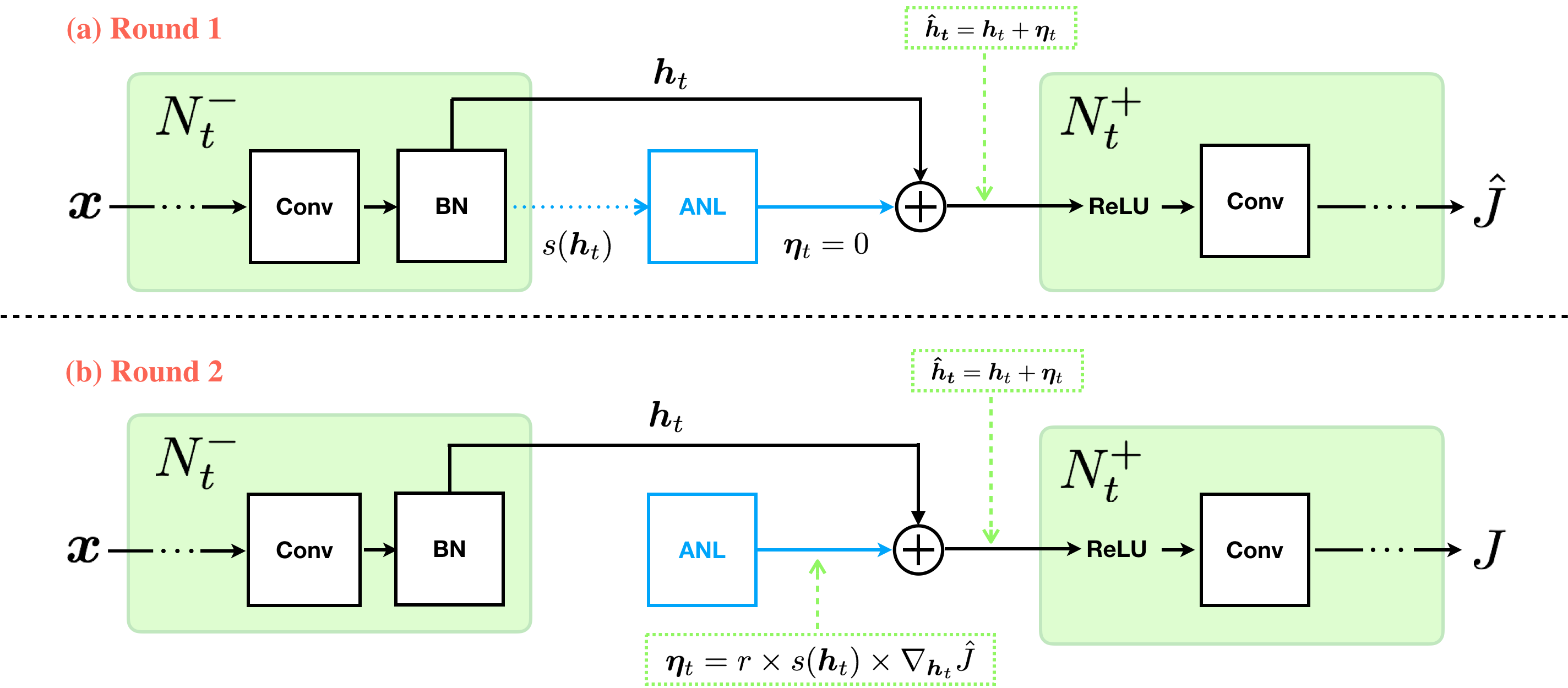}
  \caption{An illustration of two-rounds-training strategy that used by ANL. In the first round (a), ANL calculates $s(\bm h_t)$ in forward phase and $\nabla_{\bm h_t}\hat J$ in backward phase. In the second round (b), ANL will generate noise in accordance with $s(\bm h_t)$ and $\nabla_{\bm h_t}\hat J$, then the parameters of the network will be updated by back-propagation. For CANL, only one round of training is required because the gradients come from the cache.}

%  %\caption{Two rounds training strategy of the ANL algorithm. In first round (a) computes $s(\bm h_t)$ in the forward phase and $\nabla_{\bm h_t}\hat J$ in the backward phase. In second round (b), the ANL will generate noise in accordance with $s(\bm h_t)$ and $\nabla_{\bm h_t}\hat J$, then the parameters of the network will be updated by back-propagation. $N_t^-$ denotes the previous network and $N_t^+$ is the followed network.}
  \label{fig:anl}
\end{figure*}

The minimum distance to the decision boundary, which is regarded as the notation of margin, plays a foundational role in several profound theories and has empirically contributed to the overwhelming performance of both classification and regression tasks~\cite{vapnik1995nature}. It is commonly believed that a robust image classification model has a large margin. Take linear classifier into consideration, the direction of gradient vector $\nabla_{\bm x}J(\bm x, y)$ vertically points to the decision boundary, meaning that $\hat{\bm x} = \bm x + \epsilon\ \nabla_{\bm x} J(\bm x)$ is more close to the decision boundary than $\bm x$.

Taking $\hat{\bm x}$ to train will lead to higher cross-entropy loss and push the decision boundary away from $\bm x$ more significantly, which contributes to producing a larger margin classifier~\cite{goodfellow2014explaining}.

Similarly, we apply this idea to CNNs. The modern view of deep CNN architectures is that deep neural network extracts the vision features layer by layer~\cite{zeiler2014visualizing}. We use $\bm h_t$ to represent the output of $t^{th}$ layer; denote the sub-network from the first layer to the $t^{th}$ layer as $N^-_t$; denote the sub-network from $(t+1)^{th}$ layer to the last layer as $N^+_t$. $\bm h_t$ is the output of $N^-_t$ and the input of $N^+_t$.

%\hl{revise this paragraph}
%For the sub-network $N^+_t$, it is assumed that the decision boundary of the sub-network $N^+_t$ have larger margin to $\bm h_t$ through adding the specific perturbation $\epsilon \nabla_{\bm h_t}J$ into $\bm h_t$, which further improves the robustness of the models with respect to the small perturbation of $\bm h_t$.
%Therefore, adding perturbation $\epsilon \nabla_{\bm h_t}J$ instructs the sub-network $N^-_t$ to extract more distinctive features $\bm h_t$ for different class $\bm x$ to step over the margin of $N^+_t$, which thereby preventing over-fitting. Considering this, we further propose Adversarial Noise Layer (ANL). ANLs simply add the adversarial noise $\bm \eta_t$ to $\bm h_t$ and pass $\bm {\hat h_t}$ to the next layer in the process of training network.

Adding the specific perturbation $\epsilon \nabla_{\bm h_t}J$ to $\bm h_t$ leads to higher cross-entropy loss for $N^+_t$, which is also the loss for the whole network $N$. Although it is difficult to figure out the rigorous mathematical proofs on deep neural network, we conjecture that to reduce the loss, two changes will be conducted by the back-propagation update. For the sub-network $N^+_t$ which takes the $\bm h_t$ as input, the update tends to push the boundary away from $\bm h_t$. For the sub-network $N^-_t$ which generates $\bm h_t$ as output, the update tends to push the $\bm h_t$ away from the boundary. As result, the perturbation $\epsilon \nabla_{\bm h_t}J$ instructs $N^+_t$ to learn a larger margin classifier; instructs $N^-_t$ to extract more distinctive features $\bm h_t$ for different class $\bm x$ (Figure \ref{fig:feature_map}). Considering these assumptions, we further propose Adversarial Noise Layer (ANL). ANL simply adds the adversarial noise $\bm \eta_t$ to $\bm h_t$ using Eq.~(\ref{eq:ht}) and pass $\bm {\hat h_t}$ to the next layer in the process of training.
\begin{equation}
    \bm{\hat h_t} = \bm h_t + \bm \eta_t \label{eq:ht}
\end{equation}

The noise is only added during training. After the training ends, the noise layers will be wipe out and $\bm {\hat h_t} $ will be set as $\bm h_t$ in the inference phase. Therefore, ANL actually takes no extra computation in inference. According to the assumptions above, the adversarial noise $\bm \eta_t$ is designed on the basis of the gradient of $\bm h_t$ (Eq.~\ref{eq:etat}).
\begin{align}
r &= Clip_{<0, \epsilon>} \{ \mathcal N \left(\frac{\epsilon}{2}, (\frac{\epsilon}{4})^2 \right) \} \label{eq:r} \\
\bm g_t &=  \nabla_{\bm h_t} J(\bm x, y;\bm \theta, \bm \eta)|_{\bm \eta = 0} \label{eq:gt} \\
\bm \eta_t &= r\ s(\bm h_t)\ \frac{\bm g_t}{\| \bm g_t \|_\infty} \label{eq:etat}
\end{align}

%The hyper-parameter $\epsilon$ is used to control the magnitude of the noise

The random scalar $r$ is used to control the magnitude of the noise. We found that CNN models trained with dynamic magnitude noise achieve better performance than the models trained with fixed magnitude. We also multiply $r$ by $s(\bm h_t)$ for the intuition that the layer with a wide range of activations could tolerate relatively large perturbation. We verified the validity of $s(\bm h_t)$ in Experiments section.

Training with ANL requires an additional forward and backward propagation to generate the adversarial noise. We name this process as the two-rounds-training strategy which is shown in Figure \ref{fig:anl}. As the figure shows, adversarial noise is calculated after the first back-propagation and network is updated after the second back-propagation (Algorithm \ref{alg1}).

\begin{algorithm}
  \caption{Training with ANL}
  \label{alg1}
  \begin{algorithmic}
  %%\ENSURE $y = x^n$
  %%\STATE Randomly initialize network N
  \REQUIRE $\epsilon > 0$
  \REPEAT
%  	\STATE $\bm \eta \gets 0$
	\STATE Sample a batch $\{\bm X, \bm Y\}$ from the training data.
  	\STATE Forward-propagation calculate the loss $J(\bm X, \bm Y;\bm \theta, 0)$ and standard deviation of intermediate activations.
  	\STATE Backward-propagation calculate the gradients.
  	\FOR{$t$ := 1 to $L$}
  		\STATE update $\bm \eta_t$ with Eq.~(\ref{eq:etat})
  	\ENDFOR
  	\STATE Second forward-propagation get $J(\bm X, \bm Y;\bm \theta, \bm \eta)$
  	\STATE Second backward-propagation calculate the gradients.
  	\STATE Update network with $\bm \theta \gets \bm \theta - \lambda \nabla_{\bm \theta} J(\bm x, y; \bm \theta, \bm \eta)$
  \UNTIL training finish
  \end{algorithmic}
\end{algorithm}

\subsection{Class Adversarial Noise Layer}

\begin{table}[ht]
\centering
\begin{tabular}{|c|c|c|c|c|c|}
\hline
class $c$ & 0     & 1     & 2     & 3     & 4    \\ \hline
$\varphi$   & 0.77  & 0.07  & -0.13 & 0.09  & 0.03 \\ \hline
class $c$ & 5     & 6     & 7     & 8     & 9    \\ \hline
$\varphi$   & -0.10 & -0.12 & 0.05  & -0.08 & 0.02 \\ \hline
\end{tabular}
\caption{The average of gradient similarity between samples from class 0 and samples from class $c$. It should be noted that we first normalize the gradients using the maximum norm, and then calculate the similarity by Eq.~(\ref{eq:sim}).}
\label{tb:sim}
\end{table}

The two-rounds-training strategy almost doubles the training time, which is the main drawback of ANL. To reduce the extra time cost, we further proposed Class Adversarial Noise Layer (CANL) that caches gradients according to the sample category and use it for the next batch training. The noise calculation of CANL is similar to ANL's formula Eq.~(\ref{eq:etat}). The difference is that CANL obtains $\bm g_t$ from the cached gradients according to the class of the current sample. Please refer to Algorithm~\ref{alg2} for more details.

The reason why we cache gradients by class is that the gradients generated by the same class of samples are much more similar. We trained a LeNet-5~\cite{lecun1998gradient} network on MNIST for 3 epochs. Table~\ref{tb:sim} shows the average of the gradient similarity between samples from class 0 and samples from other classes. The similarity is calculated by Eq.~(\ref{eq:sim}). As shown in Table~\ref{tb:sim}, different samples from the same class produced the gradients that were much more similar than the gradients of samples from different classes.
\begin{equation}
    \varphi(\bm g_a, \bm g_b) = \dfrac{\bm g_a \cdot \bm g_b}{\Vert \bm g_a \Vert _2 \cdot \Vert \bm g_b \Vert _2} \label{eq:sim}
\end{equation}

\begin{algorithm}
  \caption{Training with CANL}
  \label{alg2}
  \begin{algorithmic}
  %%\ENSURE $y = x^n$
  %%\STATE Randomly initialize network N
  \REQUIRE $\epsilon > 0$
  \STATE $\bm G_t^c$ is the $t^{th}$ layer gradient cache for class $c$. The network containing $L$ convolutional blocks. Training data containing $C$ classes.
  \FOR{$t$ := $1$ to $L$}
  	\FOR{$c$ := $1$ to $C$}
  		\STATE $\bm G_t^c \gets 0$
  	\ENDFOR
  \ENDFOR
  \REPEAT
  	\STATE Sample a batch $\{\bm X, \bm Y\}$ from the training data.
  	\FOR{$t$ := $1$ to $L$}
  		\STATE  Get gradients $\bm g_t$ from the cache $\bm G_t$ according to $\bm Y$ to calculate noise $\bm \eta_t$.
  	\ENDFOR
  	\STATE Forward-propagation calculate the loss $J(\bm X, \bm Y;\bm \theta, \bm \eta)$
  	\STATE Backward-propagation calculate the gradients.
  	\FOR{$t$ := $1$ to $L$}
  		\STATE For every class $c$ in $\bm Y$, we random sample a corresponding $\bm x$ from $\bm X$ and then update cache with $\bm G_t^c \gets \nabla_{\bm h_t} J(\bm x, y;\bm \theta, \bm \eta)$.
  	\ENDFOR
  	\STATE Update network with $\bm \theta \gets \bm \theta - \lambda \nabla_{\bm \theta} J(\bm X, \bm Y; \bm \theta, \bm \eta)$
%  	\STATE $\bm \eta \gets 0$
%  	\STATE calculate the loss $J(\bm x,y;\bm \theta, \bm \eta)|_{\bm \eta = 0}$
%  	\STATE $\bm g_t \gets \nabla_{\bm h_t} J(\bm x, y;\bm \theta,\bm \eta)|_{\bm \eta = 0}$
%  	\FOR{$\bm h_t$ in $\bm h$}
%  		\STATE calculate $s(\bm h_t)$
%  		\STATE update $\bm \eta_t$ with Eq.~(\ref{eq:etat})
%  	\ENDFOR
%  	\STATE calculate the loss $J(\bm x,y;\bm \theta, \bm \eta)$ with adversarial noise $\bm \eta$
%  	\STATE update network with $\bm \theta \gets \bm \theta - \lambda \nabla_{\bm \theta} J(\bm x, y; \bm \theta, \bm \eta)$
  \UNTIL training finish
  \end{algorithmic}
\end{algorithm}

\subsection{Compare to LAT}

Both LAT and our approaches provide regularization by perturbing the intermediate layer activations. LAT calculates the perturbation by Eq.~(\ref{eq:lat}). 
\begin{align}
	\bm g_t^{b} &= \nabla_{h_t} J(\bm x, y;\bm \theta, \bm {\hat \eta}_{t}^{b}) \\
    \bm {\hat \eta}_{t}^{b+1} &= \epsilon\ sign(\bm g_t^{b}) \label{eq:lat}
\end{align}
$\bm g_t^b$ is the gradients of the $t^{th}$ layer, calculated by back-propagation when the model takes the batch $b$ samples as input. LAT simply stores the gradients from previous batch training and use them for the current batch training noise. $\bm {\hat \eta}_{t}^{b+1}$ is the noise calculated base on $\bm g_t^b$, and it will be added to the activations of $t^{th}$ layer during the batch $b+1$ training. LAT and our approaches differ in the following ways:

%The LAT uses the gradients from the previous batch to generate the disturbance of the current batch activation. Since the gradient similarity between different classes is relatively low, there is no guarantee that the noise calculated by the LAT will always increase the loss, which limits the regularization performance of the LAT. Unlike LAT, ANL uses the gradient of the current batch to ensure the effectiveness of the noise. On the other hand, CANL utilizes a cache gradient from the same class of samples. We demonstrated in the experimental section that ANL and CANL achieved better results than LAT in our tests.

\begin{itemize}
	\item LAT uses the gradients from the previous batch to generate perturbations for the activations of the current batch. Since the gradients similarity between different classes is relatively low, there is no guarantee that the noise calculated by LAT will always increase the loss, which limits the regularization performance of LAT. Unlike LAT, ANL adopts the gradients from the current batch to ensure the effectiveness of the noise. On the other hand, CANL takes advantage of the cached gradients from the same class samples. We demonstrate, in Experiments section, that ANL and CANL have achieved better results than LAT in our tests.
	\item We regulate noise according to the standard deviation of the layer activations. We verify the validity of standard deviation in the Experiments section. Other minor improvements have also been adopted, including random magnitude and maximum norm (Eq.~\ref{eq:etat}).
\end{itemize}

%\begin{table}[ht]
%\centering
%\caption{affect}
%\label{tb:sim}
%\begin{tabular}{|c|c|c|c|c|c|}
%\hline
%model & $\bm g$ & $\infty$ & $\bm s$     & $\mathcal N$     & accuracy    \\ \hline
%VGG   &  & &   &  & 0.09  \\ \hline
%VGG   & \ding{51} & &   &  & 0.09  \\ \hline
%VGG   & \ding{51} & \ding{51} &   &  & 0.09  \\ \hline
%VGG   & \ding{51} & \ding{51} & \ding{51}  &  & -0.08 \\ \hline
%VGG   & \ding{51} & \ding{51} & \ding{51}  & \ding{51} & -0.08 \\ \hline
%\end{tabular}
%\end{table}

\section{Experiments}
We first conduct a qualitative study on ANL and CANL by analyzing the influence of different noise on the feature maps. Then, we demonstrate the high compatibility of ANL and CANL by carrying out experiments in various mainstream CNN architectures. In the end, we validated the robustness of the models trained with ANL and CANL to the adversarial samples generated by FGSM.

% Note that it is found that the models has fully-connected layers followed with ANL perform poorly. In our work, therefore, ANL is only implemented between convolution layers.

\subsection{The Impact of Various Noise}

\begin{figure}[t]
	\centering
	\includegraphics[width=0.45\textwidth]{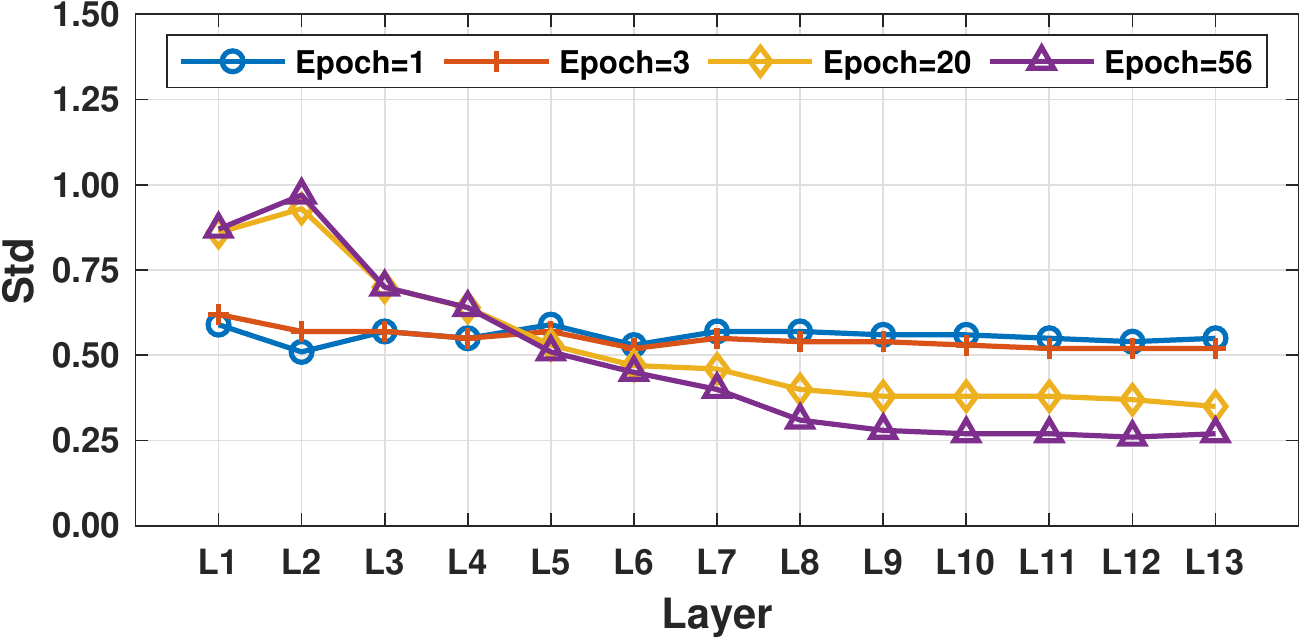}
	\caption{The standard deviation of the activation of layers under different epoch.}
  	\label{fig:std_wo_s}
\end{figure}

\begin{figure*}[t]
  \centering
  \includegraphics[width=0.8\textwidth]{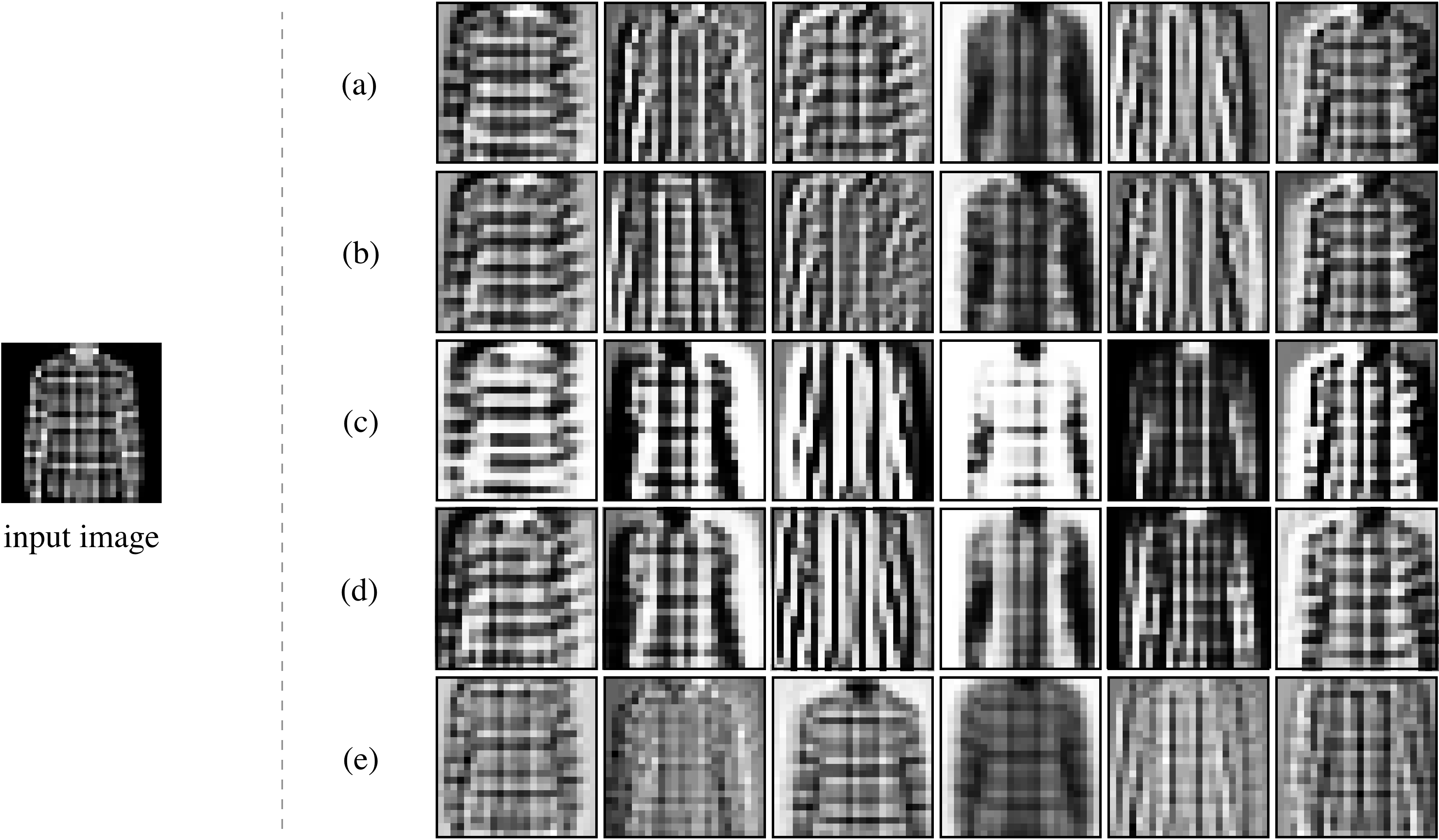}
  \caption{An illustration of feature maps from LeNet-5~\cite{lecun1998gradient} models trained with different noise. Each row represents the feature maps of all 6 channels of the first convolutional layer. The left is the input image. (a) is the feature maps from the baseline network which is trained without noise. (b) is from the network trained with the Gaussian noise from $\mathcal N(0, (\epsilon/4)^2)$ where $\epsilon=0.03$. (c) is from the network trained with ANL where $\epsilon=0.03$. (d) is from the network trained with CANL where $\epsilon=0.03$. (e) is from the network trained with ANL where $\epsilon = -0.03$.}
  \label{fig:feature_map}
\end{figure*}

\begin{table*}[ht]
\centering
\begin{tabular}{|l|l|c|c|c|c|c|c|}
\hline
Model  & Dataset       & Baseline & Gaussian & LAT  & ANL $\epsilon$ = 0.03 & ANL $\epsilon$ = -0.03 & CANL $\epsilon$ = 0.03 \\ \hline \hline
Lenet-5 & Fashion-MNIST & 9.44 & 9.49   &   -  & \textbf{8.87}                       & 12.00     &  9.21 \\ \hline
VGG-16 & CIFAR-10 & 7.81 &  7.92  &   7.35  & \textbf{6.19}                       & 54.31    &  6.32\\ \hline
ResNet-20 & CIFAR-10 & 9.71 & 9.88    &  8.90  & \textbf{7.14}                       & 74.06     & 7.45 \\ \hline
ResNet-56 & CIFAR-10 & 8.45 & 8.54    &  5.90  & \textbf{5.35}                       & 75.80     & 5.46 \\ \hline
ResNet-20 & CIFAR-100 & 36.3 & 37.2    &  33.1  & \textbf{30.5}                       & 81.27   & 31.9 \\ \hline
ResNet-56 & CIFAR-100 & 31.9 & 32.6    &  28.6  & \textbf{27.3}                       & 97.52   & 27.8 \\ \hline
\end{tabular}
\caption{The test error (\%) of models trained with various noises.}
\label{tb:noise}
\end{table*}

\begin{table*}[ht]
\centering
\begin{tabular}{|l|c|c;{2pt/2pt}c;{2pt/2pt}c;{2pt/2pt}c|c;{2pt/2pt}c;{2pt/2pt}c;{2pt/2pt}c|}
\hline
\multirow{2}{*}{Model} & \multirow{2}{*}{$\epsilon$} & \multicolumn{4}{c|}{Cifar10}                                                               & \multicolumn{4}{c|}{Cifar100}  \\ \cline{3-10}  & & \multicolumn{1}{c|}{Baseline} & \multicolumn{1}{c|}{+Dropout} & \multicolumn{1}{c|}{+ANL} & +CANL & \multicolumn{1}{c|}{Baseline} & \multicolumn{1}{c|}{+Dropout} & \multicolumn{1}{c|}{+ANL} & +CANL   \\ \hline \hline
MobileNet       &   0.02                            & 9.25                          & 9.88                      & \textbf{7.84}  &  7.98                       & 34.28                         & 34.19                           & \textbf{30.71} & 31.16\\ \hline
MobileNet v2    &   0.02                            & 7.83                          & 7.86                      & \textbf{5.54}  &    5.71                     & 29.64                         & 30.70                           & \textbf{25.51} & 25.93\\ \hline
VGG-11          &   0.05                            & 8.45                          & 8.37                      & \textbf{7.67}  &  7.79                       & 30.47                         & 29.97                           & \textbf{29.19} & 29.30\\ \hline
VGG-16          &   0.05                            & 7.47                          & 7.16                      & \textbf{5.81}  &  5.88                       & 29.45                         & 28.91                          & \textbf{26.61} & 26.73\\ \hline
ResNet-18       &   0.05                            & 5.98                          & 5.03                      & \textbf{4.21}  &  4.35                       & 24.16                         & 22.83                           & \textbf{22.23} & 22.61\\ \hline
ResNet-34       &   0.08                            & 5.34                          & 4.70                      & \textbf{3.90}  &  3.97                       & 23.42                         & 22.14                           & \textbf{21.50} & 21.82\\ \hline
PyramidNet-48   &   0.01                            & 5.62                          & 5.16                      & \textbf{4.38}  &  4.50                       & 25.14                         & 25.24                           & \textbf{23.33} & 24.03\\ \hline
PyramidNet-270  &   0.10                            & 4.68                          & 4.10                      & \textbf{3.03}  &  3.12                       & 19.55                         & 18.94 & \textbf{17.51}   & 17.94\\ \hline
\end{tabular}
\caption{Test errors (\%) with different architectures on CIFAR-10 and CIFAR-100~\cite{krizhevsky2009learning}. Baseline is the model trained without noise and dropout~\cite{srivastava2014dropout}. In (+Dropout), we insert dropout between FC layers and set drop ratio to 0.5 according to~\cite{he2016identity}. In (+ANL) and (+CANL), we insert noise layers after every batch normalization, and the choice of $\epsilon$ is based on the amount of parameters of the network.}
\label{tb:models}
\end{table*}

\begin{figure}[ht]
	\centering
	\subfigure[MobileNet-V2]{\includegraphics[width=0.4\textwidth]{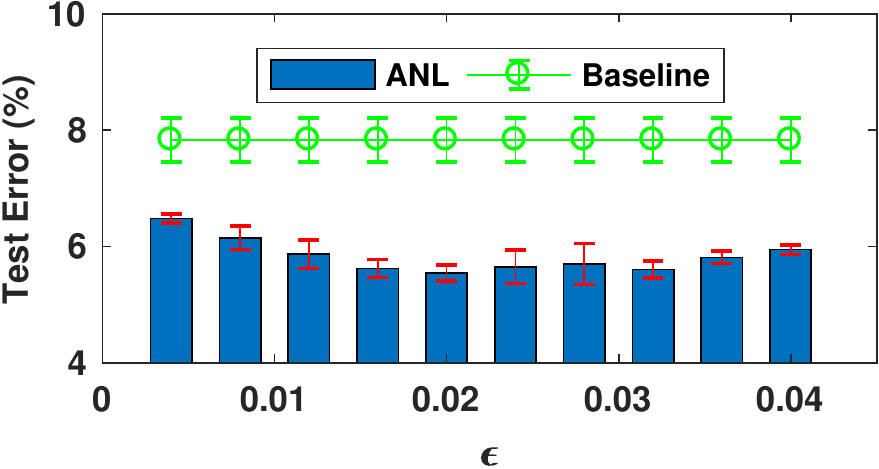}}
	\subfigure[VGG-16]{\includegraphics[width=0.4\textwidth]{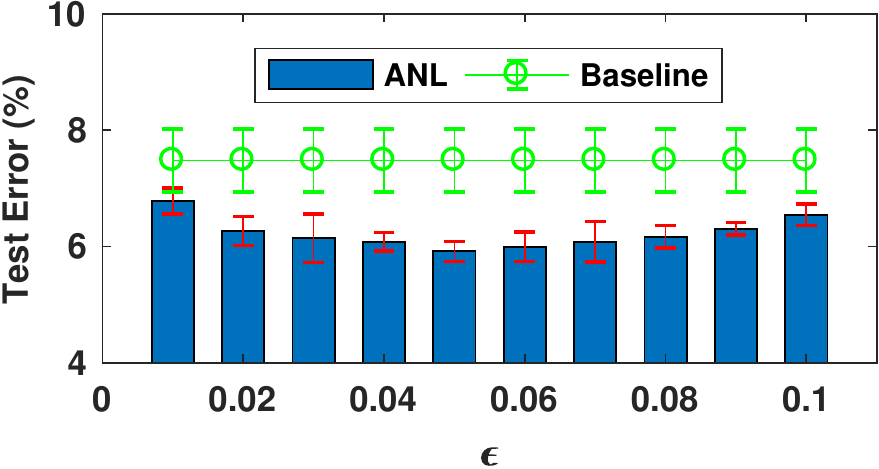}}
	\caption{The test error(\%) of (a) the MobileNet-V2~\cite{sandler2018inverted}, (b) VGG-16~\cite{simonyan2014very} on CIFAR-10, trained with ANL under different $\epsilon$. Results reported are average of 5 runs.}
  	\label{fig:epsilon}
\end{figure}

It is a straightforward way to visually illustrate the impact of adversarial noise by analyzing the feature maps calculated by the convolutional layers. We conducted the experiments with the LeNet-5~\cite{lecun1998gradient}~\footnote{CONV-\textbf{ANL}-RELU-POOL-CONV-\textbf{ANL}-RELU-POOL-FC-RELU-FC-RELU-FC} network on Fashion-MNIST~\cite{xiao2017fashion} dataset. 

Firstly, under the same initialization conditions, we trained five LeNet-5 models with different types of noise on the Fashion-MNIST dataset. Then we wipe out all of the noise layers and extract the output of the first convolution layer, and put them through a sigmoid function to get the feature maps.

Figure \ref{fig:feature_map} shows the feature maps while the input was sampled from the test dataset. Compared with the baseline, it is observed that the Gaussian noise make little difference while adversarial noise has an apparent impact on feature extraction for the network. In addition, the feature maps from the model trained with adversarial noise have sharper skeleton and structure, which indicates the model trained with adversarial noise tends to extract more distinct features for different class images. CANL shows the similar effects to ANL. It has also been found that adding the noise opposite to adversarial noise, \textit{i.e.,} setting negative $\epsilon$, tends to severely blur the feature maps, which leads to a tremendous score drop in our test. This is in line with our expectations, because according to the assumptions we discussed in the Methods section, noise that is opposite to adversarial noise might shorten the margin distance.

The quantitative comparison is shown in Table~\ref{tb:noise}. We also explore the effects of different noises on the VGG-16~\cite{simonyan2014very} and ResNet~\cite{he2016deep} models using CIFAR-10 and CIFAR-100~\cite{krizhevsky2009learning} datasets. The models are initialized with the same random seed. We follow the experiments setting employed in \cite{Sankaranarayanan18}: We use the SGD solver with Nesterov momentum of 0.9. The learning rate started at 0.1 and it is dropped by 5 every 50 epochs. All models are trained for 300 epochs. We also randomly perform horizontal flips and take a random crop with size 32x32 from images padded by 4 pixels on each side for the CIFAR-10 and CIFAR-100 experiments.

%For all the experiments, we use the SGD solver with Nesterov momentum of 0.9. The base learning rate is 0.1 and it is dropped by 5 every 60 epochs in case of CIFAR-100 and every 50 epochs in case of CIFAR-10. The total training du- ration is 300 epochs. We employ random flipping as a data augmentation procedure and standard mean/std preprocess- ing was applied conforming to the original implementations.

As shown in Table~\ref{tb:noise}, Gaussian noise slightly reduced the accuracy of the model. LAT, ANL and CANL all show regularization capabilities and achieve better accuracy than the baseline. Compared with the other regularization methods, ANL achieves the best results. CANL is slightly inferior to ANL, but significantly better than LAT. In addition, we notice that the VGG-16 model achieved 99.6\% accuracy on the CIFAR-10 training set after few epochs of training using the ANLs that configured with negative $\epsilon$, but the accuracy on test set is not higher than 45.69\%. For the ResNet models, the degeneration is even significant. We infer that the noise opposite to the adversarial noise will result in a significant reduction in model's accuracy because the model is unable to extract distinct features.

We implement ANL and CANL base on the Pytorch framework and ran experiments on the Titan Xp GPUs. The average time to train the VGG-16 baseline model on CIFAR-10 for one epoch was 14.26 seconds. The model that integrated with ANL cost 27.57 seconds to train for one epoch, while CANL took 15.83 seconds which is very close to the baseline. In general, CANL has achieved a good tradeoff between training cost and regularization capability.

\subsection{Further research on ANL and CANL capabilities}

We first verify the validity of the standard deviation in Eq.~(\ref{eq:etat}). Then we investigate the effect of the hyper-parameter $\epsilon$. Further, we demonstrate that ANL and CANL can be integrated with most of the mainstream CNN architectures and compared them with Dropout. Lastly, we conduct experiment on ImageNet to demonstrate the regularization capabilities of ANL and CANL on large-scale data set.

\subsubsection{Validity of the standard deviation}

We first train a VGG-16 network integrated with ANL on CIFAR-10, of which the adversarial noise was calculated without $s(\bm h_t)$. Figure \ref{fig:std_wo_s} shows the standard deviation of the activation of layers. As it shows, the standard deviation varies from layer to layer and changes as the model converges. The insight of using standard deviation is that if the output of the layer has a relatively large standard deviation, it should be compatible with large perturbation. The error rate drops from 6.53\% to 5.91\% by taking $s(\bm h_t)$ into account while the error rate of the baseline model that trained without noise is 7.47\%.

\subsubsection{The choice of the hyper-parameter.} $\epsilon$ is the only hyper-parameter required by ANL and CANL. To investigate the effect of different choices of $\epsilon$, we train models with two different CNN architectures on CIFAR-10 with various $\epsilon$. As the Figure~\ref{fig:epsilon} shows, each architecture has one of the most suitable choices of $\epsilon$. According to our experience, the network with more trainable parameters performs better on relative big $\epsilon$. For example, the number of parameters of MobileNet-V2 is 2.3 million, and its best $\epsilon$ is 0.02. The best $\epsilon$ for VGG-16 (14.7 million) is about 0.05, while the best $\epsilon$ for ResNet-18, which has around 11.2 million trainable parameters, is also close to 0.05. But it is not always the case, because the optimal value of $\epsilon$ is also affected by the model architecture.

\subsubsection{Integrate with various CNN architectures}

To show the compatibility of ANL and CANL in various architectures, we train various models with and without the noise layers on the CIFAR-10 and the CIFAR-100 dataset. Five architectures are adopted: MobileNet~\cite{howard2017mobilenets}, MobileNet v2~\cite{sandler2018inverted}, VGG~\cite{simonyan2014very}, ResNet~\cite{he2016deep} and PyramidNet~\cite{DPRN}. We also compared the proposed algorithms with dropout.

The learning rate started at 0.1 and would be divided by 2 unless the last 5 epoch validation loss reaches lower value than the best ever seen. Each learning rate stays at least 5 epoch and the minimum learning rate is 0.001. We use a weight decay of 5e-4 and momentum of 0.9 for all experiments. We train the baseline for 200 epoch. In comparison, the network is trained with ANL and CANL for the first 100 epoch. Then we disable the noise layers and set the learning rate to the minimum (\textit{i.e.} 1e-3) before we train the network for another 100 epoch.

The following observations can be made from Table~\ref{tb:models}. The proposed noise layer is compatible with various CNN architectures and shows a noticeable improvement from baseline. ANL had achieved the best results in all comparisons, while CANL presented very close regularization capability to ANL. For the experiments on the MobileNet model, Dropout slightly reduced accuracy, but ANL and CANL still demonstrated good regularization.

\subsubsection{Imagenet Experiment} In order to test the applicability of the proposed regularization methods on large-scale dataset, we conduct an experiment on the ImageNet~\cite{ILSVRC15} dataset. We implement AlexNet~\cite{krizhevsky2012imagenet} integrated with CANL, and set $\epsilon=0.05$. Both the baseline and the regularized model are trained from scratch to 90 epochs. The learning rate started from 0.01 decayed by 10 every 30 epochs. The classification accuracy of the baseline is 56.35\%, while the accuracy of the regularized model is 60.52\%. This shows that our approaches can significantly improve the performance of deep neural networks on the large datasets like ImageNet.

\subsection{Adversarial Attack Evaluation}

\begin{table}[t]
\centering
\begin{tabular}{|r|c|c|c|c|c|c|}
\hline
$\delta$           & 0 & 2 & 4 & 6 & 8  \\ \hline \hline
Baseline    & 92.0  &  53.8    &  42.2    &  37.4    &   33.8      \\ \hline
AT($\epsilon$=0.05) & 91.9 & 74.5 & 68.8 & 59.7 & 53.9 \\ \hline
ANL($\epsilon$=0.01)  &  92.6 &   67.1   &   57.0   &   52.4   &  47.7          \\ \hline
ANL($\epsilon$=0.05) & \textbf{93.9} &   \textbf{79.2}    &    \textbf{71.8}  &   \textbf{63.7}   &  \textbf{61.0}        \\ \hline
CANL($\epsilon$=0.01) & 92.6  &  66.9    &   56.3   &   51.0   &  46.3          \\ \hline
CANL($\epsilon$=0.05) & 93.7  & 75.7  &  65.0 & 58.3 & 55.8        \\ \hline
\end{tabular}
\caption{The test accuracy(\%) of the VGG-16 networks on the adversarial examples that generated by FGSM with different $\delta$.}
\label{tb:FGSM}
\end{table}

%  Comparison of classification accuracy (%) between the different variants of the proposed approach and FGS method (FGS-orig) for different values of ε

Deep convolutional neural networks are easily fooled by careful designed adversarial examples. Plenty of literatures~\cite{goodfellow2014explaining,papernot2017practical,kurakin2016adversarial,moosavi2017universal} show that small perturbations cause well-designed deep networks to misclassify the image easily. Although the main focus of our proposed methods is to prevent model from overfitting, we have found it is helpful in improving the robustness of the model to the adversarial examples as well.

In our experiment, we use the Fast Gradient Sign Method (FGSM)~\cite{goodfellow2014explaining} to generate adversarial examples. Assuming the input image $\bm x$ is in the range 0 to 255, the perturbed image $\bm{\hat{x}}$ is generated as $\bm{\hat{x}} = \bm x + \delta\ sign(\nabla_{\bm{x}} J(\bm x))$. The value of $\delta$ is usually set to small number relative to 255 to generate the perturbations imperceptible to human but degrade the accuracy of a network significantly.

We train six different VGG-16 networks on CIFAR-10 for the FGSM white-box attack tests. We add the noise layers not only after the convolution layers but also to the input $\bm x$. The baseline model is trained without noise. For the ANL and CANL tests, we try different values of $\epsilon$. We also test the AT algorithm introduced by \cite{goodfellow2014explaining}, which enhances the robustness of the network by using the adversarial examples for training.

The results is shown in Table~\ref{tb:FGSM}. The value of $\delta$ indicate the strength of the adversarial perturbation, generated by FGSM, which is added to the image to produce adversarial examples. From the table, we have the following observations: (1) $\delta=0$ means no adversarial perturbation is added to the images. Compared to AT, the proposed methods significantly improve the performance of the original data. (2) ANL and CANL show comparable robustness enhancements to AT (3) For ANL and CANL, a larger $\epsilon$ increases the perturbation to the intermediate activations, providing a stronger robustness enhancement for the network.

\section{Conclusion}

In this paper, we propose two regularization algorithms called "Adversarial Noise Layer" and "Class Adversarial Noise Layer". They are easy to implement and can be integrated with the most of CNN-based models. The proposed methods require only one hyper-parameter $\epsilon$, and the choice of $\epsilon$ is related to the number of trainable parameters of the network. The noise is only added during the training process, so there is no additional computation cost for CNN models in the inference phase. Models trained with ANL or CANL have been proved to be more robust under the FGSM attack.

Currently, we only apply ANL and CANL in image classification tasks. In future work, we will explore other computer vision tasks such as object detection and face recognition. It is also interesting to explore the application of ANL and CANL in the area beyond computer vision, such as natural language processing and voice recognition.

{
    \bibliographystyle{aaai}
    \bibliography{ref}

\begin{thebibliography}{}

\bibitem[\protect\citeauthoryear{Goodfellow, Shlens, and
  Szegedy}{2015}]{goodfellow2014explaining}
Goodfellow, I.~J.; Shlens, J.; and Szegedy, C.
\newblock 2015.
\newblock Explaining and harnessing adversarial examples.
\newblock In {\em International Conference on Learning Representations (ICLR)}.

\bibitem[\protect\citeauthoryear{Han, Kim, and Kim}{2017}]{DPRN}
Han, D.; Kim, J.; and Kim, J.
\newblock 2017.
\newblock Deep pyramidal residual networks.
\newblock In {\em {IEEE} Conference on Computer Vision and Pattern Recognition
  (CVPR)},  6307--6315.

\bibitem[\protect\citeauthoryear{He \bgroup et al\mbox.\egroup
  }{2016a}]{he2016deep}
He, K.; Zhang, X.; Ren, S.; and Sun, J.
\newblock 2016a.
\newblock Deep residual learning for image recognition.
\newblock In {\em {IEEE} Conference on Computer Vision and Pattern Recognition
  (CVPR)},  770--778.

\bibitem[\protect\citeauthoryear{He \bgroup et al\mbox.\egroup
  }{2016b}]{he2016identity}
He, K.; Zhang, X.; Ren, S.; and Sun, J.
\newblock 2016b.
\newblock Identity mappings in deep residual networks.
\newblock In {\em European Conference on Computer Vision (ECCV)},  630--645.

\bibitem[\protect\citeauthoryear{He \bgroup et al\mbox.\egroup
  }{2017}]{DBLP:conf/iccv/HeGDG17}
He, K.; Gkioxari, G.; Doll{\'{a}}r, P.; and Girshick, R.~B.
\newblock 2017.
\newblock Mask {R-CNN}.
\newblock In {\em International Conference on Computer Vision (ICCV)},
  2980--2988.

\bibitem[\protect\citeauthoryear{Howard \bgroup et al\mbox.\egroup
  }{2017}]{howard2017mobilenets}
Howard, A.~G.; Zhu, M.; Chen, B.; Kalenichenko, D.; Wang, W.; Weyand, T.;
  Andreetto, M.; and Adam, H.
\newblock 2017.
\newblock Mobilenets: Efficient convolutional neural networks for mobile vision
  applications.
\newblock {\em arXiv preprint arXiv:1704.04861}.

\bibitem[\protect\citeauthoryear{Krizhevsky and
  Hinton}{2009}]{krizhevsky2009learning}
Krizhevsky, A., and Hinton, G.
\newblock 2009.
\newblock Learning multiple layers of features from tiny images.
\newblock In {\em https://www.cs.toronto.edu/\~{}kriz/cifar.html}.

\bibitem[\protect\citeauthoryear{Krizhevsky, Sutskever, and
  Hinton}{2012}]{krizhevsky2012imagenet}
Krizhevsky, A.; Sutskever, I.; and Hinton, G.~E.
\newblock 2012.
\newblock Imagenet classification with deep convolutional neural networks.
\newblock In {\em Conference on Neural Information Processing Systems (NIPS)},
  1106--1114.

\bibitem[\protect\citeauthoryear{Kurakin, Goodfellow, and
  Bengio}{2017}]{kurakin2016adversarial}
Kurakin, A.; Goodfellow, I.; and Bengio, S.
\newblock 2017.
\newblock Adversarial machine learning at scale.
\newblock {\em International Conference on Learning Representations (ICLR)}.

\bibitem[\protect\citeauthoryear{LeCun \bgroup et al\mbox.\egroup
  }{1998}]{lecun1998gradient}
LeCun, Y.; Bottou, L.; Bengio, Y.; and Haffner, P.
\newblock 1998.
\newblock Gradient-based learning applied to document recognition.
\newblock {\em Proceedings of the IEEE} 86(11):2278--2324.

\bibitem[\protect\citeauthoryear{Lee \bgroup et al\mbox.\egroup
  }{2015}]{DBLP:conf/aistats/LeeXGZT15}
Lee, C.; Xie, S.; Gallagher, P.~W.; Zhang, Z.; and Tu, Z.
\newblock 2015.
\newblock Deeply-supervised nets.
\newblock In {\em Proceedings of the Eighteenth International Conference on
  Artificial Intelligence and Statistics (AISTATS)},  562--570.

\bibitem[\protect\citeauthoryear{Moosavi{-}Dezfooli \bgroup et al\mbox.\egroup
  }{2017}]{moosavi2017universal}
Moosavi{-}Dezfooli, S.; Fawzi, A.; Fawzi, O.; and Frossard, P.
\newblock 2017.
\newblock Universal adversarial perturbations.
\newblock In {\em {IEEE} Conference on Computer Vision and Pattern Recognition
  (CVPR)},  86--94.

\bibitem[\protect\citeauthoryear{Papernot \bgroup et al\mbox.\egroup
  }{2017}]{papernot2017practical}
Papernot, N.; McDaniel, P.~D.; Goodfellow, I.~J.; Jha, S.; Celik, Z.~B.; and
  Swami, A.
\newblock 2017.
\newblock Practical black-box attacks against machine learning.
\newblock In {\em Proceedings of the 2017 {ACM} on Asia Conference on Computer
  and Communications Security (ASIACCS)},  506--519.

\bibitem[\protect\citeauthoryear{Qian \bgroup et al\mbox.\egroup
  }{2017}]{DBLP:journals/corr/abs-1711-10098}
Qian, R.; Tan, R.~T.; Yang, W.; Su, J.; and Liu, J.
\newblock 2017.
\newblock Attentive generative adversarial network for raindrop removal from a
  single image.
\newblock {\em arXiv preprint arXiv:1711.10098}.

\bibitem[\protect\citeauthoryear{Rosenblatt}{1957}]{wang2017origin}
Rosenblatt, F.
\newblock 1957.
\newblock {\em The perceptron, a perceiving and recognizing automaton Project
  Para}.
\newblock Cornell Aeronautical Laboratory.

\bibitem[\protect\citeauthoryear{Russakovsky \bgroup et al\mbox.\egroup
  }{2015a}]{ILSVRC15}
Russakovsky, O.; Deng, J.; Su, H.; Krause, J.; Satheesh, S.; Ma, S.; Huang, Z.;
  Karpathy, A.; Khosla, A.; Bernstein, M.; Berg, A.~C.; and Fei-Fei, L.
\newblock 2015a.
\newblock {ImageNet Large Scale Visual Recognition Challenge}.
\newblock {\em International Journal of Computer Vision (IJCV)}
  115(3):211--252.

\bibitem[\protect\citeauthoryear{Russakovsky \bgroup et al\mbox.\egroup
  }{2015b}]{DBLP:journals/ijcv/RussakovskyDSKS15}
Russakovsky, O.; Deng, J.; Su, H.; Krause, J.; Satheesh, S.; Ma, S.; Huang, Z.;
  Karpathy, A.; Khosla, A.; Bernstein, M.~S.; Berg, A.~C.; and Li, F.
\newblock 2015b.
\newblock Imagenet large scale visual recognition challenge.
\newblock {\em International Journal of Computer Vision (IJCV)}
  115(3):211--252.

\bibitem[\protect\citeauthoryear{Sandler \bgroup et al\mbox.\egroup
  }{2018}]{sandler2018inverted}
Sandler, M.; Howard, A.; Zhu, M.; Zhmoginov, A.; and Chen, L.-C.
\newblock 2018.
\newblock Inverted residuals and linear bottlenecks: Mobile networks for
  classification, detection and segmentation.
\newblock {\em arXiv preprint arXiv:1801.04381}.

\bibitem[\protect\citeauthoryear{Sankaranarayanan \bgroup et al\mbox.\egroup
  }{2018}]{Sankaranarayanan18}
Sankaranarayanan, S.; Jain, A.; Chellappa, R.; and Lim, S.
\newblock 2018.
\newblock Regularizing deep networks using efficient layerwise adversarial
  training.
\newblock In {\em Proceedings of the Thirty-Second {AAAI} Conference on
  Artificial Intelligence},  4008--4015.
\newblock {AAAI} Press.

\bibitem[\protect\citeauthoryear{Simonyan and
  Zisserman}{2015}]{simonyan2014very}
Simonyan, K., and Zisserman, A.
\newblock 2015.
\newblock Very deep convolutional networks for large-scale image recognition.
\newblock In {\em International Conference on Learning Representations (ICLR)}.

\bibitem[\protect\citeauthoryear{Srivastava \bgroup et al\mbox.\egroup
  }{2014}]{srivastava2014dropout}
Srivastava, N.; Hinton, G.; Krizhevsky, A.; Sutskever, I.; and Salakhutdinov,
  R.
\newblock 2014.
\newblock Dropout: A simple way to prevent neural networks from overfitting.
\newblock {\em Journal of Machine Learning Research (JMLR)} 15(1):1929--1958.

\bibitem[\protect\citeauthoryear{Vapnik}{1995}]{vapnik1995nature}
Vapnik, V.~N.
\newblock 1995.
\newblock {\em The Nature of Statistical Learning Theory}.
\newblock Springer science \& business media.

\bibitem[\protect\citeauthoryear{Wan \bgroup et al\mbox.\egroup
  }{2013}]{wan2013regularization}
Wan, L.; Zeiler, M.; Zhang, S.; Le~Cun, Y.; and Fergus, R.
\newblock 2013.
\newblock Regularization of neural networks using dropconnect.
\newblock In {\em International Conference on Machine Learning (ICML)},
  1058--1066.

\bibitem[\protect\citeauthoryear{Xiao, Rasul, and
  Vollgraf}{2017}]{xiao2017fashion}
Xiao, H.; Rasul, K.; and Vollgraf, R.
\newblock 2017.
\newblock Fashion-mnist: a novel image dataset for benchmarking machine
  learning algorithms.
\newblock {\em arXiv preprint arXiv:1708.07747}.

\bibitem[\protect\citeauthoryear{Xie \bgroup et al\mbox.\egroup
  }{2016}]{xie2016disturblabel}
Xie, L.; Wang, J.; Wei, Z.; Wang, M.; and Tian, Q.
\newblock 2016.
\newblock Disturblabel: Regularizing cnn on the loss layer.
\newblock In {\em {IEEE} Conference on Computer Vision and Pattern Recognition
  (CVPR)},  4753--4762.

\bibitem[\protect\citeauthoryear{Zeiler and
  Fergus}{2013}]{zeiler2013stochastic}
Zeiler, M.~D., and Fergus, R.
\newblock 2013.
\newblock Stochastic pooling for regularization of deep convolutional neural
  networks.
\newblock In {\em International Conference on Learning Representations (ICLR)}.

\bibitem[\protect\citeauthoryear{Zeiler and
  Fergus}{2014}]{zeiler2014visualizing}
Zeiler, M.~D., and Fergus, R.
\newblock 2014.
\newblock Visualizing and understanding convolutional networks.
\newblock In {\em European Conference on Computer Vision (ECCV)},  818--833.

\bibitem[\protect\citeauthoryear{Zhong \bgroup et al\mbox.\egroup
  }{2017}]{zhong2017random}
Zhong, Z.; Zheng, L.; Kang, G.; Li, S.; and Yang, Y.
\newblock 2017.
\newblock Random erasing data augmentation.
\newblock {\em arXiv preprint arXiv:1708.04896}.

\end{thebibliography}
}

\end{document}